\documentclass[conference]{IEEEtran}
\usepackage{times}

\usepackage[numbers]{natbib}
\usepackage{multicol}
\usepackage[bookmarks=true]{hyperref}

\usepackage[pdftex]{graphicx}
\usepackage[cmex10]{amsmath}
\usepackage{amssymb}
\usepackage{import}
\usepackage{bm}

\usepackage{color}
\usepackage{units}
\usepackage{url}
\usepackage{float}
\usepackage{epstopdf}
\usepackage{stmaryrd}

\usepackage{units}
\usepackage{siunitx}

\usepackage{mathtools}

\newcommand{\half}{\ensuremath{\frac{1}{2}}}
\newcommand{\norm}[1]{\left\| #1 \right\|}

\newcommand{\mvec}[1]{\boldsymbol{#1}}

\newcommand{\mmat}[1]{\begin{bmatrix} #1 \end{bmatrix} }

\newcommand{\msum}[3]{\displaystyle\sum\limits_{#1}^{#2} {#3}}

\newcommand{\mrb}[1]{\left( #1 \right)} %enclose in round brackets
\newcommand{\msb}[1]{\left[ #1 \right]} %enclose in square brackets
\newcommand{\commentOut}[1]{}

\newcommand{\figref}[1]{Fig.~\ref{#1}}
\newcommand{\secref}[1]{Section~\ref{#1}}

\newcommand{\realNums}{\ensuremath{\mathbb{R}}}
\newcommand{\skewSym}[1]{\ensuremath{\,S\!\mrb{#1}}}
\newcommand{\skewSymInv}[1]{\ensuremath{\,{v}\!\mrb{#1}}}

\newcommand{\trace}[1]{\ensuremath{\mathrm{tr}\!\mrb{#1}}}
\newcommand{\diag}[1]{\ensuremath{\mathrm{diag}\!\mrb{#1}}}

\newcommand{\SOthree}{\ensuremath{\mathrm{SO}\mrb{3}}}
\newcommand{\sothree}{\ensuremath{\mathrm{so}\mrb{3}}}

\newcommand{\ddt}{\ensuremath{\frac{\mathrm{d}}{\mathrm{d}t}}}
\newcommand{\ddtSq}{\ensuremath{\frac{\mathrm{d}^2}{\mathrm{d}t^2}}}

\newcommand{\identityMat }{\ensuremath{\mvec{I}}}

\newcommand{\rotMat}{\ensuremath{\mvec{R}}}
\newcommand{\rotAngle}{\ensuremath{\rho}}
\newcommand{\rotAxis}{\ensuremath{\mvec{n}}}

\newcommand{\rotAngleErr}{\ensuremath{\rho}_\mathrm{e}}
\newcommand{\rotAxisErr}{\ensuremath{\mvec{n}_\mathrm{e}}}

\newcommand{\rotMatDes}{\ensuremath{\mvec{R}_\mathrm{des}}}
\newcommand{\rotMatErr}{\ensuremath{\mvec{R}_\mathrm{e}}}

\newcommand{\rotMatReduced}{\ensuremath{\mvec{R}_r}}
\newcommand{\rotMatYaw}{\ensuremath{\mvec{R}_y}}
\newcommand{\rotAngleReduced}{\ensuremath{\rho_r}}
\newcommand{\rotAxisReduced}{\ensuremath{\mvec{n}_r}}
\newcommand{\rotAngleYaw}{\ensuremath{\rho_y}}
\newcommand{\rotAxisYaw}{\ensuremath{\mvec{n}_y}}

\newcommand{\baseVec}[1]{\ensuremath{\mvec{e}_{#1}}}

\newcommand{\angVel}{\ensuremath{\mvec{\omega}}}
\newcommand{\angVelDes}{\ensuremath{\mvec{\omega}_\mathrm{des}}}
\newcommand{\angVelErr}{\ensuremath{\mvec{\omega}_\mathrm{e}}}

\newcommand{\angAcc}{\ensuremath{\mvec{\alpha}}}
\newcommand{\angAccDes}{\ensuremath{\mvec{\alpha}_\mathrm{des}}}
\newcommand{\angAccErr}{\ensuremath{\mvec{\alpha}_\mathrm{e}}}

\newcommand{\pos}{\ensuremath{\mvec{p}}}
\newcommand{\posDes}{\ensuremath{\mvec{p}_\mathrm{des}}}
\newcommand{\translAccDes}{\ensuremath{\mvec{a}_\mathrm{des}}}

\newcommand{\translAcc}{\ensuremath{\mvec{a}}}
\newcommand{\thrustDir}{\ensuremath{\baseVec{3}}}
\newcommand{\gravity}{\ensuremath{\mvec{g}}}
\newcommand{\mass}{\ensuremath{m}}
\newcommand{\thrustMag}{\ensuremath{f_\Sigma}}
\newcommand{\thrustMagDes}{\ensuremath{f_{\Sigma,\mathrm{des}}}}
\newcommand{\mmoi}{\ensuremath{\mvec{J}}}
\newcommand{\moments}{\ensuremath{\mvec{\tau}}}

\newcommand{\angAccInpSS}{\ensuremath{\mvec{\alpha}_\mathrm{e,des}^\mathrm{SS}}}
\newcommand{\angAccInpRV}{\ensuremath{\mvec{\alpha}_\mathrm{e,des}^\mathrm{RV}}}
\newcommand{\angAccInpTP}{\ensuremath{\mvec{\alpha}_\mathrm{e,des}^\mathrm{QTP}}}
\newcommand{\angAccInpNEW}{\ensuremath{\mvec{\alpha}_\mathrm{e,des}^\mathrm{New}}}

\newcommand{\lyapFnSS}{\ensuremath{J^\mathrm{SS}}}
\newcommand{\lyapFnRV}{\ensuremath{J^\mathrm{RV}}}

\newcommand{\lyapFnNEW}{\ensuremath{J^\mathrm{New}}}

\newcommand{\gainAtt}{\ensuremath{K_{R}}}
\newcommand{\gainRates}{\ensuremath{K_\omega}}

\newcommand{\gainPos}{\ensuremath{k_{p}}}
\newcommand{\gainVel}{\ensuremath{k_{\dot{p}}}}

\newcommand{\gainAttRed}{\ensuremath{k_{r}}}
\newcommand{\gainAttYaw}{\ensuremath{k_{y}}}

\title{\LARGE \bf
Multicopter attitude control for recovery from large disturbances
}

\author{
Mark W. Mueller% <-this % stops a space
\thanks{The author is with the Mechanical Engineering Department, UC Berkeley.
\newline        {\tt\small mwm@berkeley.edu}}%
}

\begin{document}

\maketitle
\thispagestyle{empty}
\pagestyle{empty}

%%%%%%%%%%%%%%%%%%%%%%%%%%%%%%%%%%%%%%%%%%%%%%%%%%%%%%%%%%%%%%%%%%%%%%%%%%%%%%%%
\begin{abstract}

We present a novel, high-performance attitude control law for multicopters, with a view to recovery from large disturbances. The controller is compared to three well-established alternatives from the literature. All controllers considered are identical to first order, but differ in their computation of the attitude error. We show that the popular use of the skew-symmetric part of the rotation matrix is problematic from a safety perspective, and specifically that the closed loop system may linger at large attitude errors for an arbitrary duration (leading to potential failures of practical systems). The novel proposed controller prioritizes the error in the vehicle thrust direction, and is shown to outperform a similar, existing controller from the literature. Stability follows via a Lyapunov function, and the controller is validated in experiments. This novel controller is especially attractive in safety-critical situations, where a multicopter may be required to recover from large initial disturbances.

\end{abstract}

\section{Introduction}
\label{secIntro}

A crucial requirement for successful control of a multicopter UAV is the control of its attitude, or orientation. 
The design of typical multicopters means that they are able to produce a torque in an arbitrary direction, so that the attitude dynamics are fully actuated.
The practical need for good multicopter attitude control is furthermore complemented by the intriguing and elegant nature of the nonlinear dynamics of orientation, which has led to a large number of publications on the topic.

An excellent introduction to attitude control is given in \cite{chaturvedi2011rigid}, which provides a detailed discussion on the properties of orientations and their dynamics, and proposes some control laws while providing in-depth stability analyses. 
Specifically, a major argument therein is to use the rotation matrix directly for control, rather than (for example) the Euler symmetric parameters / quaternion of rotation. 
An example of the use of attitude control using quaternions is given in \cite{fresk2013full}.
%Indeed, there are a \todo{multitude} of works describing attitude control using quaternions, for example \cite{fresk2013full} \todo{More quaternions!}.
Alternative strategies may use, for example, the Euler angles \cite{lupashin2014platform}, which are intuitive to describe but have undesirable properties at large orientations.

The agility of multicopters is undisputed, and they are capable of remarkable feats (e.g. \cite{mueller2011quadrocopter,mellinger2012trajectory,ritz2012cooperative,mueller2015relaxed,falanga2017aggressive}). 
As a result, they perform an increasingly large set of tasks in daily life, including inspection, surveillance, transport of goods, and performing as part of theater groups. 
As part of this increasing ubiquity, they are expected to encounter (and recover from) an ever larger set of potential disturbances.

%This work is motivated by the search for a control strategy that allows a multicopter to recover well from large attitude disturbances, such as may be experienced after impact with a foreign object. 
The goal of this paper is as follows. 
First, we briefly present three popular multicopter attitude control strategies, and discuss their relative advantages and disadvantages.
These controllers differ only in how attitude error affects the commanded angular acceleration, specifically including the use of the skew-symmetric component of the rotation matrix (as in \cite{lee2010geometric}), the rotation vector (axis-angle of rotation, as in \cite{bullo1995proportional,yu2015high}), or a quaternion-based tilt prioritization (as in \cite{brescianini2013nonlinear}). 
Specifically, we argue that the skew-symmetric control strategy, though shown to have almost-global stability properties, in fact represents a safety concern when used in practical systems due to the system potentially dwelling arbitrarily long at attitudes near $180^\circ$ away from the desired.
Then, inspired by \cite{brescianini2013nonlinear}, we present a novel tilt-prioritizing attitude controller for multicopters, which prioritizes the ability of the multicopter to achieve a target acceleration. 
This novel controller is analyzed using the rotation matrix, and rotation vector, allowing for stability analysis using a particularly simple Lyapunov function. 
Numerical results are given, comparing the performance of the various controllers, and the highlighting the advantages of the proposed control law. 
The contribution of this paper is thus the derivation of a novel tilt-prioritizing attitude control law, the comparison of this with popular attitude approaches from the literature, a demonstration and discussion of safety concerns of a popular and widely used attitude controller, and numerical as well as experimental validation of the control law.

It should be noted that, in addition to \cite{brescianini2013nonlinear}, other forms of tilt prioritization have been used. 
For example, in \cite{yu2015high}, a similar attitude decomposition is used as in the proposed method; however the prioritization is done by dividing the recovery trajectory into two segments: the first controlling a tilt angle, and the second subsequently controlling a yaw-like angle.

We proceed by briefly describing the salient features of multicopter dynamics and the mathematics of attitudes, and how the vehicle attitude influences its motion. 
In \secref{secControllers} we describe the controllers from the literature and derive the novel controller. 
Numerical examples illustrate the properties of the controllers in \secref{secPerformance}, experiments are given in \secref{secExpValidation} and we conclude in \secref{secConclusion}. 
 
\section{Problem formulation}
\label{secDynamics}

Conventional multicopters are characterized by having an even number (at least four) of equally-sized fixed-pitch propellers arranged in a rotationally symmetric pattern about a geometric center which approximately coincides with the vehicle battery, electronics, and payload. 
The propellers are arranged in alternating handedness, so that their aerodynamic reaction torques can be made to sum to zero. 

Beyond this typical configuration, many other designs are possible and have been considered. 
Examples include using propellers of vastly different diameters for increased efficiency \cite{pounds2002design}, having the propellers be tiltable \cite{ryll2012modeling}, and a vehicle with propellers unaligned so that its translation is fully actuated \cite{mehmood2016maneuverability}.
Though such vehicles do not exactly conform to the below description, their attitude control problems are similar in that they are able to produce arbitrary 3D torques, so that their attitude dynamics is fully actuated.
%There are, however, multicopter designs where the attitude dynamics are not fully actuated, see e.g. \cite{mueller2015relaxed,zhang2016controllable}.

\subsection{Dynamics}

\begin{figure}
  \centering
  \includegraphics{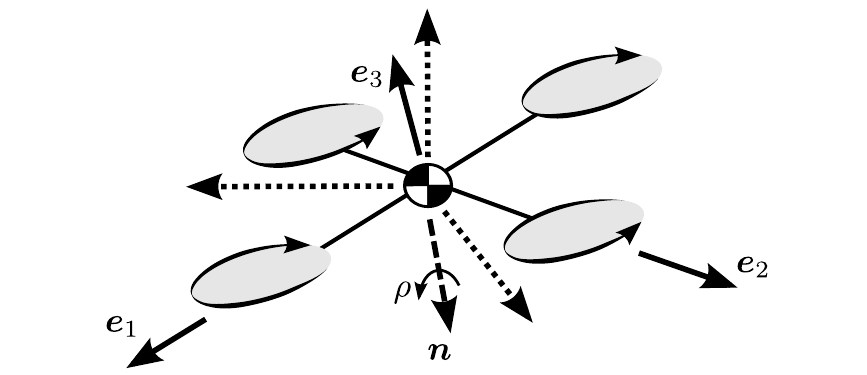}
  \caption{
  A multicopter with body-fixed axes $\baseVec{1}$, $\baseVec{2}$, and $\baseVec{3}$, rotated by $\rotMat$ from the desired axes represented by dotted lines. 
  The rotation $\rotMat$ is about the unit vector $\rotAxis$, by the angle $\rotAngle$.
  }
  \label{figModel}
\end{figure}

The orientation of the multicopter (relating a body-fixed coordinate system to an inertial coordinate system) is described by the rotation matrix $\rotMat\in\SOthree$, while the angular velocity is given by $\angVel\in\realNums^3$. 
The rotation matrix affects the motion of the multicopter primarily through the orientation of the thrust vector, which has fixed direction relative to the vehicle body.

The vehicle's propellers all produce thrust in the same body fixed direction $\thrustDir$, and with scalar magnitude $\thrustMag$, shown schematically in \figref{figModel}.
The vehicle has a mass $\mass$, and is acted upon by gravitational acceleration $\gravity$, so that the translational acceleration $\translAcc$ of the vehicle is given by 
\begin{align}
	\translAcc = \frac{1}{\mass} \rotMat \thrustDir \thrustMag + \gravity. \label{eqDynamicsTranslAcc}
\end{align}
Thus, by controlling the attitude of the vehicle and specifying the total thrust force, the translational acceleration can be controlled. 
Notable from \eqref{eqDynamicsTranslAcc} is that only two of the attitude's three degrees of freedom are relevant to the translational dynamics, with specifically the rotation about the vehicle's $\baseVec{3}$ axis being unimportant. 

The attitude evolves as 
\begin{align}
	\dot{\rotMat} = \rotMat \skewSym{\angVel}
\end{align}
where $\skewSym{\cdot}: \realNums^3\rightarrow\sothree$ produces the skew-symmetric matrix form of the vector argument (often called the ``hat-map''), specifically if $\mvec{x}=\mrb{x_1, x_2, x_3}$ then 
\begin{align}
	\skewSym{\mvec{x}} = \mmat{0 & -x_3 & x_2\\ x_3 & 0 & -x_1 \\ -x_2 & x_1 & 0}
\end{align}
Notably, for $\mvec{x},\mvec{y}\in\realNums^3$ and $\rotMat\in\SOthree$ \cite{bernstein2009matrix}
\begin{align}
  \skewSym{\mvec{x}} &= -\skewSym{\mvec{x}}^T
\\\skewSym{\mvec{x}}\mvec{y} &= \mvec{x}\times\mvec{y} = -\skewSym{\mvec{y}}\mvec{x}
\\\skewSym{\rotMat\mvec{x}} &= \rotMat\skewSym{\mvec{x}} \rotMat^T
\end{align}
The inverse of the above is the function $\skewSymInv{\cdot}:\sothree\rightarrow\realNums^3$, so that
\begin{align}
	\skewSymInv{\skewSym{\mvec{x}}} = \mvec{x}
\end{align}

The angular acceleration $\angAcc$ is a function of the vehicle's mass moment of inertia tensor $\mmoi$, the external moments acting on the vehicle $\moments$, and the current angular velocity, as 
\begin{align}
	\angAcc = \dot{\angVel} = \mmoi^{-1}\mrb{\moments - \skewSym{\angVel}\mmoi\angVel} \label{eqDynamicsAngAcc}
\end{align}

The configuration of all conventional multicopters (quad-, hexa-, and octocopters) is such that the vehicle can produce an arbitrary (up to motor force saturation) three-dimensional moment $\moments$ independent of the total force $\thrustMag$.
The computation of the moment and total force as a function of the individual propeller forces follows in a straight-forward manner from the vehicle's geometry and the properties of the propellers. 
An important feature of typical multicopters is that they are able to produce much larger torques in directions perpendicular to the thrust vector than they can about the thrust vector.
This is due to the large distance that the propellers are from the center of mass, which may be more than an order of magnitude greater than the propellers' aerodynamic torque-from-thrust ratio. 
For an in-depth discussion on computing forces for agile maneuvering, see for example \cite{faessler2017thrust}. 

\subsection{Control problem}
Thus, from \eqref{eqDynamicsAngAcc} it can be seen that an arbitrary angular acceleration $\angAcc$ may be produced at any instantaneous angular velocity (up to motor force saturation). 
This motivates the use of the angular acceleration as control input for the attitude subsystem, and specifically means that a multicopter attitude may be considered as fully actuated, giving the simpler attitude dynamics:
\begin{align}
	\dot{\rotMat} &= \rotMat\skewSym{\angVel}
\\	\dot{\angVel} &= \angAcc
\end{align}

We consider the control problem as that of controlling the vehicle's orientation to a desired attitude $\rotMatDes$, which has an associated desired angular velocity $\angVelDes$ and angular acceleration $\angAccDes$ so that
\begin{align}
	\ddt\rotMatDes =& \rotMatDes \skewSym{\angVelDes}
\\  \ddt \angVelDes =& \angAccDes
\end{align}
The rotation error $\rotMatErr$ and its angular velocity $\angVelErr$ are defined as 
\begin{align}
	\rotMatErr =& \rotMatDes^{-1} \rotMat \label{eqDefineRotMatErr}
\\  \ddt \rotMatErr =& \rotMatErr \skewSym{\angVelErr} \label{eqDefineAngVelErr}
\end{align}

Substituting the definition of the rotation error, and after some algebra, it follows that 
\begin{align}
  \angVelErr &= \angVel - \rotMatErr^{-1} \angVelDes
\\ \angAccErr &= \ddt \angVelErr = \angAcc - \rotMatErr^{-1} \angAccDes + \skewSym{\angVelErr} \angVelDes \label{eqDefAngAccErr}
\end{align}

For the sake of compactness, we will use $\angAccErr$ as the control input, noting that the command torque is recovered by substituting \eqref{eqDefAngAccErr} into \eqref{eqDynamicsAngAcc}.

For analysis, it is often more intuitive to express a rotation matrix $\rotMat$ as a rotation vector, decomposed into an angle $\rotAngle\in[0,\pi]$ and a unit-length rotation axis $\rotAxis$ (often called the eigen-axis).
The relationship between these quantities is given by \cite{shuster1993survey}
\begin{align}
	\rotMat = \cos\rotAngle\identityMat + \mrb{1-\cos\rotAngle}\rotAxis\rotAxis^T + \sin\rotAngle\skewSym{\rotAxis}
	\label{eqRotMatFromAxisAngle}
\end{align}
From this, the angle and axis for a given rotation matrix may also be straight-forwardly recovered, except for rotations by $180^\circ$, when rotations about $\rotAxis$ and $-\rotAxis$ are equivalent, and zero rotations where the axis is irrelevant.

\section{Controllers}
\label{secControllers}

We consider four different control layouts, three of which have a history of application in interesting and challenging environments, while the fourth is a novel algorithm.
All controllers have an action proportional to the angular velocity, and a component of action proportional to (some representation) of the vehicle attitude.
The main difference in the controllers is the representation of this attitude, and though the controllers are identical to first order it will be shown in the following section that important differences emerge for large attitude errors. 

\subsection{Skew-symmetric control}

This control is as described in \cite{lee2010geometric}, and is presented first due to its especially wide use in the literature (with some examples including \cite{sreenath2013geometric,goodarzi2013geometric,lee2013nonlinear,simha2017almost,rashad2017design}) and specifically its use in the influential \cite{mahony2012aerial}.
Note that we use a substantially different notation and representation in the hope of offering a unified comparison and additional insight.

The attitude error is computed from the skew-symmetric component of the rotation matrix, so that the desired angular acceleration is given by
\begin{align}
	\angAccInpSS := -\gainRates \angVelErr - \half \gainAtt \skewSymInv{\rotMatErr-\rotMatErr^T} \label{eqDefInputSS}
\end{align}
with $\gainRates$ and $\gainAtt$ positive definite controller gains, each in $\realNums^{3\times3}$.
Note that the attitude component may be rewritten, via \eqref{eqRotMatFromAxisAngle}, as 
\begin{align}
	\half \skewSymInv{\rotMatErr-\rotMatErr^T} = \sin \rotAngleErr\, \rotAxisErr
\end{align}
so that
\begin{align}
	\angAccInpSS = -\gainRates \angVelErr - \gainAtt \sin\rotAngleErr\, \rotAxisErr.\label{eqSkewSymmControllerAxAngle}
\end{align}

The stability of this controller can be investigated using the below Lyapunov function
\begin{align}
	\lyapFnSS := \frac{1}{2} \angVelErr^T \gainAtt^{-1} \angVelErr + \frac{3 - \trace{\rotMatErr}}{2}
\end{align}
where the trace is related to the rotation angle as follows
\begin{align}
	\trace{\rotMatErr} =& \msum{i=1}{3}{\baseVec{i}^T\rotMatErr\baseVec{i}} \label{eqRotTraceBaseVectors}
\\ =& 2\cos{\rotAngleErr} + 1
\end{align}
with the last equality following from \eqref{eqRotMatFromAxisAngle}. 
Thus, the Lyapunov function can be written more intuitively as 
\begin{align}
	\lyapFnSS := \half \angVelErr^T \gainAtt^{-1} \angVelErr + \mrb{1-\cos\rotAngleErr}
\end{align}
which can be easily verified to be a valid candidate Lyapunov function. 

The time derivative of the trace follows from \eqref{eqRotTraceBaseVectors} as 
\begin{align}
	\ddt \trace{\rotMatErr} =& 
	%\msum{i=1}{3}{\baseVec{i}^T  \rotMatErr \skewSym{\angVelErr} \baseVec{i}} =
	 -\msum{i=1}{3}{\baseVec{i}^T  \rotMatErr \skewSym{\baseVec{i}} \angVelErr}.
\end{align}
Furthermore, by direct computation, it can be shown that 
%see ``test.py'' script
\begin{align}
  \msum{i=1}{3}{\baseVec{i}^T  \rotMatErr \skewSym{\baseVec{i}} } = \skewSymInv{\rotMatErr-\rotMatErr^T}^T
\end{align}

Therefore, taking the time derivative of the Lyapunov function yields 
\begin{align}
	\ddt \lyapFnSS =& \angVelErr^T\mrb{\gainAtt^{-1} \angAcc + \half \skewSymInv{\rotMatErr-\rotMatErr^T}}
%\\=& \angVelErr^T\mrb{-\gainAtt^{-1}\gainRates \angVelErr - \mrb{\gainAtt^{-1} \gainAtt-\identityMat} \skewSymInv{\rotMatErr-\rotMatErr^T} }
\\=& -\angVelErr^T \gainAtt^{-1}\gainRates \angVelErr \leq 0
\end{align}
Asymptotic stability follows by noting that the second time derivative is given as 
\begin{align}
\begin{split}
	\ddtSq \lyapFnSS=& \angVelErr^T \mrb{\gainAtt^{-1}\gainRates + \gainRates^T\gainAtt^{-T}}\cdot
	\\ & \mrb{-\gainAtt \sin\rotAngleErr\rotAxisErr - \gainRates \angVelErr }
\end{split}
\end{align}
The negative semi-definite derivative implies that $\lyapFnSS(t)\leq\lyapFnSS(0)$; in turn bounding the angular velocity $\angVelErr$.
From this bound, the second derivative is bounded, so that $\ddt\lyapFnSS$ is uniformly continuous and integrable. 
Thus  as $t\rightarrow\infty$, by Barbalat's lemma, $\ddt\lyapFnSS \rightarrow 0$ and specifically $\angVelErr\rightarrow0$. 
Substituting the control law allows to conclude that the orientation error thus also converges to identity if $\rotAngleErr\neq\pi$, establishing asymptotic stability.

%\todo{Not allowed to LaSalle: So that asymptotic stability follows by invoking LaSalle's principle \cite{sastry2013nonlinear}.}
%\todo{KS: Control in [9] provides exponential stability for a set of rotation errors. Do you want to specialize your development to illustrate this?}

A very closely related control strategy weights each term in the sum on the right-hand side of \eqref{eqRotTraceBaseVectors} with distinct scalar values \cite{chaturvedi2011rigid}.
The resulting closed-loop system has some desirable properties (and its stability is proven in \cite{chaturvedi2011rigid} without relying on a Lyapunov function), and introduces three saddle points into the attitude dynamics not present in \eqref{eqDefInputSS}. 
That controller is however similar enough that we only consider the simpler form wherein all terms are equally weighted.

\subsection{Rotation vector control}
For this control strategy, the attitude error is computed using the rotation vector (so that the attitude part of the feedback is proportional to the angle error).
The desired angular acceleration is 
\begin{align}
	\angAccInpRV := -\gainRates \angVelErr - \gainAtt \rotAngleErr \, \rotAxisErr \label{eqDefInputRotVec}
\end{align}
where $\gainRates$ and $\gainAtt$ are again positive definite gain matrices.
Note that the angle is bounded by definition so that $0\leq\rotAngleErr\leq\pi$.
Note, furthermore, the similarity to \eqref{eqDefInputSS}, with the crucial difference being the use of the angle, rather than its sine.

Stability of the resulting closed-loop system is analyzed using the Lyapunov function
\begin{align}
	\lyapFnRV :=& \half \angVelErr^T \gainAtt^{-1} \angVelErr + \half \mrb{\arccos\frac{\trace{\rotMatErr}-1}{2}}^2 \label{eqLyapRotVec}
\end{align}
which may be simplified as 
\begin{align}
	\lyapFnRV =  \half \angVelErr^T \gainAtt^{-1} \angVelErr + \half \rotAngleErr^2
\end{align}

Noting that the time derivative of the rotation angle is given by \cite[(270)]{shuster1993survey}
\begin{align}
	\ddt \rotAngleErr = \rotAxisErr^T \angVelErr \label{eqTimeDerivativeRotAngle}
\end{align}
the time derivative of \eqref{eqLyapRotVec} follows as 
\begin{align}
	\ddt \lyapFnRV =& \angVelErr^T \gainAtt^{-1} \angAcc + \rotAngleErr \angVelErr^T\rotAxisErr
	\\ =& - \angVelErr^T \gainAtt^{-1} \gainRates \angVelErr  \leq 0
\end{align}
From this point, asymptotic stability follows similarly as for the skew-symmetric controller by noting the bounded second derivative and invoking Barbalat's lemma. 
Again, this requires that $\rotAngleErr\neq\pi$.
%\todo{Not allowed to use LaSalle's principle allowing to conclude asymptotic stability.}
%\todo{Also, talk about the advantages\ldots}
%\todo{KS: Do you have exponential stability for $\rotAngle<\pi/2$?}

The use of the rotation vector is conceptually elegant, as the control action is proportional to the angle, even for large attitude errors. 
This means that the system's closed loop behavior (if restricted to a single rotary degree of freedom) will behave like a second-order damped system even for large angles, as long as $\rotAngleErr<180^\circ$.  
There is, however, a discontinuity in the control input as the angle crosses `through' $180^\circ$, as the sign flips for $\rotAxisErr$.

\subsection{Quaternion-based tilt-prioritized control}
This controller is based on the intuition that the most important part of a quadcopter's attitude is the orientation of its thrust axis, and thus prioritizes controlling this direction above the single other attitude degree of freedom. 
The controller is presented in \cite{brescianini2013nonlinear}, and that derivation bases on the quaternion of rotation.
This is here translated to the rotation matrix representation, to better place in context with the two other presented methods, and as a preview of the proposed controller. 

The attitude error is divided into two parts: a prioritized `reduced' attitude $\rotMatReduced$, representing the shortest rotation which would align the thrust direction with the desired thrust direction, and a rotation about the thrust axis $\rotMatYaw$ representing the remaining rotation about the thrust axis.
Specifically, the reduced attitude is the smallest rotation for which 
\begin{align}
	\rotMatErr\rotMatReduced^T  \thrustDir =& \thrustDir \label{eqDefRedAtt1}
\end{align}
and 
\begin{align}
  \rotMatErr  =& \rotMatYaw \rotMatReduced \label{eqDefRedAtt2}
\end{align}
from which follows that $\rotMatYaw$ is a rotation purely about $\thrustDir$.
Note that this angle is equivalent, to first order, to the Euler `yaw' angle (of the 3-2-1 yaw-pitch-roll sequence).

The axis $\rotAxisReduced$ and angle $\rotAngleReduced$ corresponding to $\rotMatReduced$ may be computed as 
\begin{align}
  \rotAngleReduced =& \arccos {\thrustDir^T\rotMatErr\thrustDir} \label{eqComputeReducedRotAngle}
\\\rotAxisReduced =& \frac{\skewSym{\rotMatErr^T\thrustDir}\thrustDir}{\norm{\skewSym{\rotMatErr^T\thrustDir}\thrustDir}} = \frac{\skewSym{\rotMatErr^T\thrustDir}\thrustDir}{\sin\rotAngleReduced} \label{eqComputeReducedRotAxis}
\end{align}

Note that, by construction, $\rotAxisReduced$ is perpendicular to the thrust direction. 
Also note that the potential division by zero in \eqref{eqComputeReducedRotAxis} is of no concern, since the corresponding angle of \eqref{eqComputeReducedRotAngle} is then zero (and thus an arbitrary axis may be specified).

The yaw rotation axis $\rotAxisYaw$ and angle $\rotAngleYaw$ are computed as those corresponding to the rotation matrix $\rotMatYaw$, which follows as 
\begin{align}
  \rotMatYaw =& \rotMatErr\rotMatReduced^{-1}   
\end{align}
wherein $\rotAxisYaw$ will always be either parallel or anti-parallel to $\thrustDir$. 
Because the axes of rotation for $\rotMatReduced$ and $\rotMatYaw$ are perpendicular, the angles are related as below \cite[(114)]{shuster1993survey}
\begin{align}
	\cos\mrb{\half\rotAngleErr} = \cos\mrb{\half\rotAngleReduced}\cos\mrb{\half\rotAngleYaw} \label{eqRelationErrorAngleReducedYaw}
\end{align}
so that $\rotAngleReduced \leq \rotAngleErr$ (due to the monotonicity of $\cos$ for angles in $\msb{0,\half\pi}$).
%\todo{KS: also mention relation between $\rotAngleError$ and $\rotAngle_r$, if any.}

The control action is given as below, where the use of half-angles follows from the original rotation-quaternion-based form \cite{brescianini2013nonlinear}:
\begin{align}
	\angAccInpTP = -\gainRates \angVelErr - 2 \gainAttRed \rotAxisReduced \sin\frac{\rotAngleReduced}{2} - 2 \gainAttYaw \rotAxisYaw \sin\frac{\rotAngleYaw}{2}
\label{eqDefInputQTP}
\end{align}
and specifically $\gainAttRed>\gainAttYaw$ to prioritize reducing the tilt error $\rotAngleReduced$.
The stability proof of this controller is somewhat involved, and will not be repeated here -- the reader is referred to \cite{brescianini2013nonlinear}. 

As infinitesimal rotations commute (so that their rotation-vector-representations may be added for composition of the related rotations), it may be seen that this control law linearizes in the same fashion as the prior controllers. 

Like the rotation-vector control, the control action is discontinuous at $\rotAngleErr=180^\circ$.
However, unlike the rotation-vector based controller, the nonlinearity of the sin function means that large pure rotations will not behave like second-order damped rotations. 

\subsection{Proportional tilt-prioritized control}
Inspired by the rotation-vector and quaternion-based tilt-prioritizing controllers, we propose an alternative form wherein the control action is proportional to the angle, rather than the $\sin$ of the half-angle as imposed by the quaternion formulation. 
This has four advantages: (1) like the rotation-vector controller, the closed-loop system response is exactly like a second-order damped system for any initial rotation about one of the principle axes, 
(2) the Lyapunov function is substantially simpler,
(3) as the relative priority of the reduced attitude error approaches that of the overall attitude, the controller converges to the rotation-vector controller, and
(4) the proposed controller outperforms the quaternion-based controller both in speed of convergence and efficiency of control action.

Using again $\gainAttRed$ as the gain applied to the tilt angle error, and $\gainAttYaw$ the gain for the remaining angle, we use the control law
\begin{align}
\angAccInpNEW = - \gainRates \angVelErr - \gainAttYaw \rotAngleErr \, \rotAxisErr - \mrb{\gainAttRed - \gainAttYaw} \rotAngleReduced \rotAxisReduced \label{eqDefInputNew}
\end{align}
wherein, as before, $\rotAngleErr$ refers to the total attitude error, and $\rotAngleReduced$ to the reduced attitude error.
Unlike the quaternion-based controller, this does not require the computation of $\rotAngleYaw$ and $\rotAxisYaw$.
Note that the controller has two terms related to attitude error, and that for symmetric control ($\gainAttYaw=\gainAttRed$) it reduces to the rotation vector control \eqref{eqDefInputRotVec}.  

Stability follows via the following Lyapunov function
\begin{align}
  \lyapFnNEW := \half \angVelErr^T \angVelErr + \half {\gainAttYaw} \rotAngleErr^2 + \half \mrb{\gainAttRed - \gainAttYaw} \rotAngleReduced^2
\end{align}
This is positive definite for $0<\gainAttYaw\leq \gainAttRed$ (in other words, the control for the tilt angle must at least as important as control of the overall attitude).

The derivative of the reduced tilt angle error $\rotAngleReduced$ can be computed from \eqref{eqComputeReducedRotAngle}-\eqref{eqComputeReducedRotAxis} as
\begin{align}
 \ddt \rotAngleReduced &= \rotAxisReduced^T \angVelErr
\end{align}
(which is similar \eqref{eqTimeDerivativeRotAngle}), so that 
\begin{align}
  \ddt \lyapFnNEW =& \angVelErr^T \mrb{\angAcc + \gainAttYaw \rotAngleErr \, \rotAxisErr + \mrb{\gainAttRed - \gainAttYaw} \rotAngleReduced \rotAxisReduced}
\\ =& - \angVelErr^T \gainRates \angVelErr \leq 0
\end{align}
The application of Barbalat's lemma (as for the skew-symmetric controller) allows to conclude asymptotic stability for $\rotAngleErr\neq\pi$, where it should be noted that no additional constraint on $\rotAngleReduced$ is required since, by \eqref{eqRelationErrorAngleReducedYaw}, $\rotAngleReduced\leq\rotAngleErr$.
Note that, again, this controller performs the same as all others to first order.

In the limit, as $\gainAttYaw$ approaches zero, the controller tends to only control the vehicle's tilt angle, a behavior shared with the quaternion-based tilt-prioritizing control. 
However, a useful additional property of this controller is that it smoothly converges to the rotation-vector based control as the relative importance of the tilt angle is reduced (i.e. as $\gainAttYaw\rightarrow \gainAttRed$).
This means that a designer may implement this control law, even if it is not needed to have a large increase in tilt angle stiffness.
 
\section{Performance}
\label{secPerformance}

In this section, the controllers are compared in a few conditions that highlight different properties. 
As noted, all controllers behave similarly for small errors (and setting $\gainAtt=\diag{\gainAttRed,\gainAttRed,\gainAttYaw}$).
However, for larger attitude errors, salient differences appear. 
Specifically, it will be shown that the skew-symmetric controller will maintain an almost 180$^\circ$ attitude error for an arbitrary duration; this is a particular concern when deploying in safety-critical environments. 

The rotation-vector based controller performs best with respect to the total attitude error in each case; and specifically the rotation-vector based controller appears to be substantially preferable to the skew-symmetric control.
However, the tilt-prioritizing controllers (both the quaternion-based controller as well as the proposed controller) outperform the rotation-vector controller when considering only the thrust direction error. 
Moreover, the proposed controller will be shown to be superior to the quaternion-based controller. 

\subsection{Common simulation parameters}
For all simulations, the attitude control parameters are as follows:
\begin{align}
	\gainAtt    &= \diag{\gainAttRed,\gainAttRed,\gainAttYaw}
\\  \gainAttRed &= 4 \ \mathrm{s}^{-2}
\\  \gainAttYaw &= 1 \ \mathrm{s}^{-2}
\\  \gainRates  &= \sqrt{2}\;\diag{2, 2, 1} \ \mathrm{s}^{-1}
\end{align}
Thus all controllers behave, to first order, as mass-spring-dampers with a natural frequency of $2$rad/s in the tilt direction, and $1$rad/s in the yaw direction, and a damping ratio of $\sqrt{1/2}\approx0.707$.
All controllers share the same control parameters for the experiments.

\subsection{Arbitrarily slow convergence with the skew-symmetric controller}
\label{secPerfInitAngVel}

\begin{figure}
  \centering
  \includegraphics{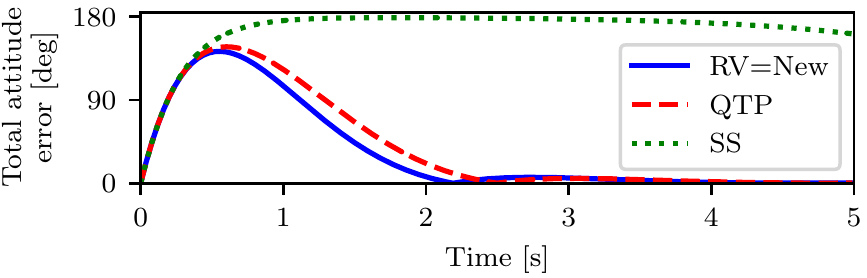}
  \caption{
  Example from \secref{secPerfInitAngVel}: recovering from a large initial angular velocity, showing catastrophic performance of the skew-symmetric controller. 
  	`SS' refers to the skew-symmetric controller \eqref{eqDefInputSS}, `RV' to the rotation-vector controller \eqref{eqDefInputRotVec}, `QTP' to the quaternion-based tilt-prioritizing controller \eqref{eqDefInputQTP}, and `New' to the proposed controller \eqref{eqDefInputNew}.
  	The behavior of the rotation-vector based controller is identical to that of the new controller.
  }
  \label{figCaseLargeInitVel}
\end{figure}

Consider a vehicle starting at rest, but with an attitude error of $\rho\approx180^\circ$.
From \eqref{eqSkewSymmControllerAxAngle} it is clear that the angular acceleration commanded by the skew-symmetric controller is approximately zero, due to taking the sine of the angle. 
In flight, this presents itself as a safety concern -- though the attitude will eventually converge to the desired attitude as long as it is not at exactly $\rho=180^\circ$, this may take an arbitrarily long time. 
Specifically, the `stiffness' of the attitude control starts \emph{decreasing} as the attitude error exceeds $90^\circ$.
Of the four controllers, this is unique to the skew symmetric controller. 

To illustrate this potential safety concern more vividly, consider a vehicle with zero initial attitude error, but with angular velocity $\angVel(0)=\mrb{10.8,0,0}$rad/s.
The response of the system is shown in \figref{figCaseLargeInitVel}.
Notable is that all controllers rapidly bring the angular velocity to zero, however the skew symmetric controller's attitude error lingers near 180$^\circ$.
The rotation-vector based controller and the proposed new controller perform identically, and slightly outperform the quaternion-based thrust prioritizing controller. 

The potential for slow convergence of the skew-symmetric controller has been previously noted \cite{lee2012exponential}, but the controller remains popular in the literature (e.g. \cite{simha2017almost,sreenath2013geometric,rashad2017design}).

\subsection{The advantage of tilt-prioritization}
\label{secPerfAdvantageTiltPrior}
\begin{figure}
  \centering
  \includegraphics{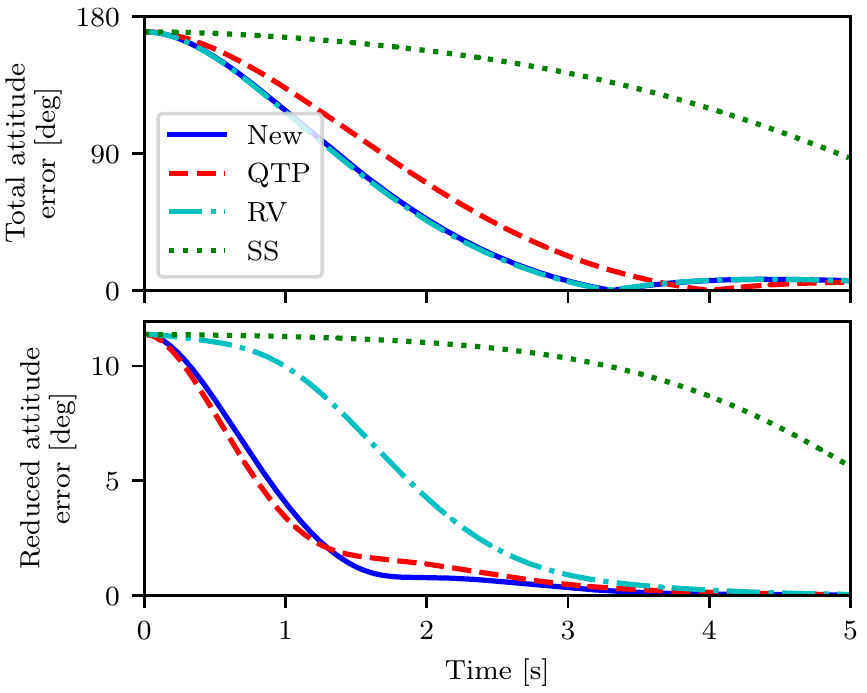}
  \caption{
  Example from \secref{secPerfAdvantageTiltPrior}: recovering from a large initial yaw error, with small initial tilt error, showing the more rapid convergence of the thrust direction for the proposed controller.
  }
  \label{figCaseLargeLargeYaw}
\end{figure}

By design, the tilt-prioritizing controllers should cause a faster convergence of the vehicle's thrust direction to the desired thrust direction. 
As an example of this behaviour, consider a vehicle starting at rest, but rotated by 170$^\circ$ about the axis $\rotAxis(0)\approx\mrb{0.0995,0,0.995}$, so that the vehicle has a large yaw error, but only a slight tilt error.
The performance of the controllers is compared in \figref{figCaseLargeLargeYaw}, where it can be seen that both the quaternion-based and proposed controller reduce the tilt error much faster than the rotation-vector based control. 
Furthermore, in this specific case, the performance of the proposed controller is practically indistinguishable from the rotation-vector based controller for the total attitude error, while the quaternion-based tilt-prioritizing controller performs notably poorer. 

The skew-symmetric controller again performs particularly poorly. 
Note that this is, in some sense, a ``friendlier'' initial condition than the previous example, since the vehicle now primarily has an initial yaw error, which ideally should have little effect on the vehicle's dynamics.
This kind of error, furthermore, is relatively typical on takeoff (e.g. if an operator places the vehicle incorrectly) due to the visual rotational symmetry typical of multicopters.
The yaw error, moreover, is not particularly carefully chosen; for $10/180\approx5\%$ of the yaw range the performance will be no better than that shown in the figure.

\subsection{Disadvantage of tilt prioritization}
\label{secPerfDisadvantageTiltPrior}
\begin{figure}
  \centering
  \includegraphics{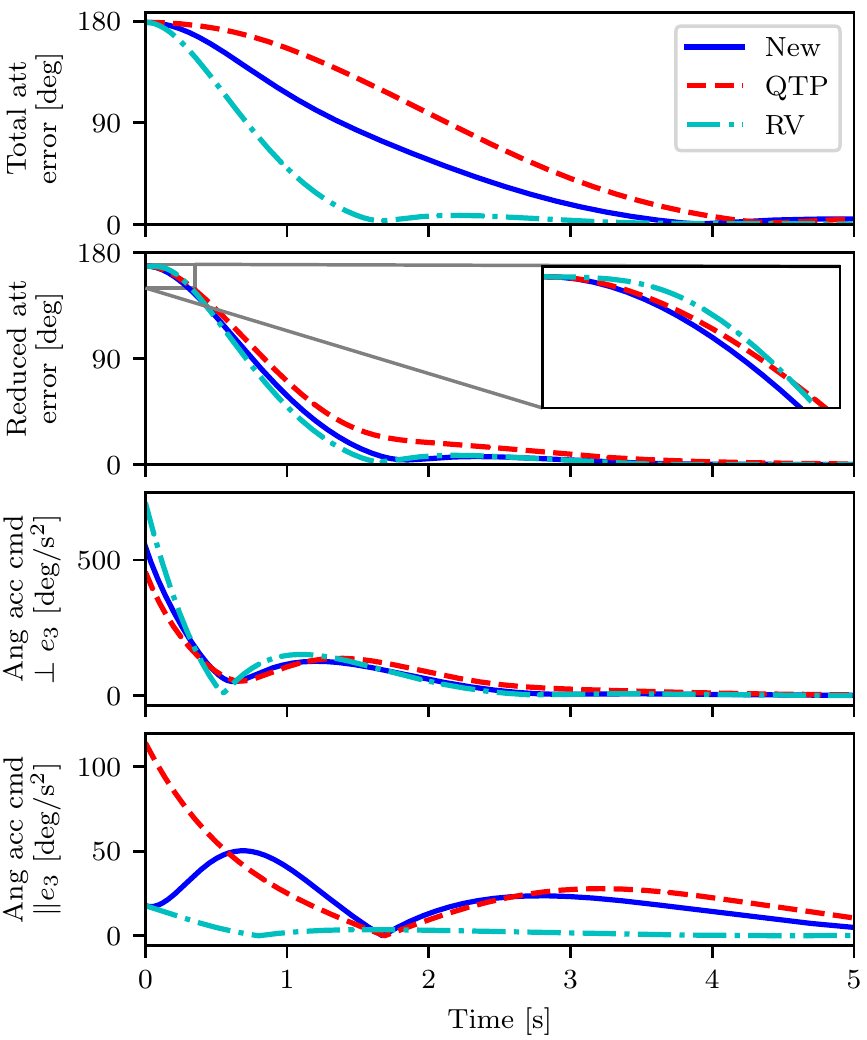}
  \caption{
  Example from \secref{secPerfDisadvantageTiltPrior}: recovering from a large initial tilt error, with small initial yaw error, showing an example where the rotation-vector based control outperforms the proposed thrust-prioritizing controller.
  From top to bottom: the total attitude error $\rotAngle$; the reduced attitude error $\rotAngleReduced$; the magnitude of the component of commanded angular acceleration perpendicular to the thrust direction $\thrustDir$; and the magnitude of the component of commanded angular acceleration parallel to the thrust direction.
  	The inset shows a detail of the reduced attitude response.
  }
  \label{figCaseLargeLargeTilt}
\end{figure}

For the total attitude error (rather than only the tilt error), the rotation-vector based control will generally outperform the proposed controller, as it acts directly only on this error.
Consider, for example, an initial rotation of 179$^\circ$ about the axis $\rotAxis(0)\approx\mrb{0.995,0,0.0995}$.
A rotation about any axis perpendicular to the vehicle's thrust direction will end with the vehicle having approximately zero tilt error, however, by varying the choice of rotation axis the remaining yaw error can be either zero, or as large as $180^\circ$. 

\figref{figCaseLargeLargeTilt} compares the performance of the controllers. 
The rotation-vector based control, as expected, performs best when considering the overall attitude error, and in this case the tilt error also ultimately converges fastest.
Both the quaternion-based and newly proposed  tilt-prioritizing controllers perform worse for overall attitude error, but do initially outperform the rotation-vector based controller for the tilt angle. 
Notable is that reduced attitude error is reduced most quickly by the proposed controller initially. 

Notable, moreover, is that the proposed controller outperforms the quaternion-based controller both for tilt error, and for total angle error. 
Furthermore, it does this while commanding a lower peak angular acceleration about the vehicle's thrust axis, as is shown at the bottom of \figref{figCaseLargeLargeTilt}, even though the total attitude error decreases more rapidly. 
This property is generally desirable in multicopters, as they are able to produce much larger torques about $\baseVec{1}$ and $\baseVec{2}$ than about the thrust direction $\baseVec{3}$, and large angular acceleration commands about $\baseVec{3}$ are likely to quickly cause saturation of the motor forces.

\section{Experimental validation}
\label{secExpValidation}

We present two experiments to demonstrate the proposed controller, and also to emphasize that the identified issues discussed in the previous section are not limited to carefully constructed numerical simulations.
Experiments are run using a Crazyflie 2.0 quadcopter, operating in an indoor motion capture space. 
The first experiment demonstrates a vehicle taking off with a large initial yaw error.
In the second experiment large disturbances are simulated, to demonstrate the recovery of a vehicle from large initial errors.

The vehicle is controlled with a simple cascaded control structure, where the desired translational acceleration is computed with a proportional-derivative controller on the position error, 
\begin{align}
  \translAccDes := -\gainPos \mrb{\pos - \posDes} - \gainVel \dot{\pos}
\end{align}
From the desired acceleration, the desired orientation is generated as the smallest rotation matrix $\rotMatDes$ for which 
\begin{align}
	\translAccDes = \frac{1}{\mass} \rotMatDes \thrustDir \thrustMagDes + \gravity
\end{align}
where $\thrustMagDes$ is the desired thrust magnitude, also defined through the above. 

Note that this is an unsophisticated control structure, however it suffices to demonstrate the proposed control law.

\commentOut{
Specifically, in the experiments, the following constants are used:
\begin{align}
  & \gainPos = 2 \mathrm{s}^{-2}
  & \gainVel = 4\sqrt{2}\mathrm{s}^{-1}
\\&\gainAttRed = 167 \mathrm{s}^{-2} 
  &\gainAttYaw = 2 \mathrm{s}^{-1} 
\\&\gainAtt    = \diag{\gainAttRed,\gainAttRed,\gainAttYaw}
  &\gainRates  = \diag{33,33,2} \mathrm{s}^{-1}
\end{align}
}

\begin{figure}
  \centering
  \includegraphics[width=0.7\linewidth]{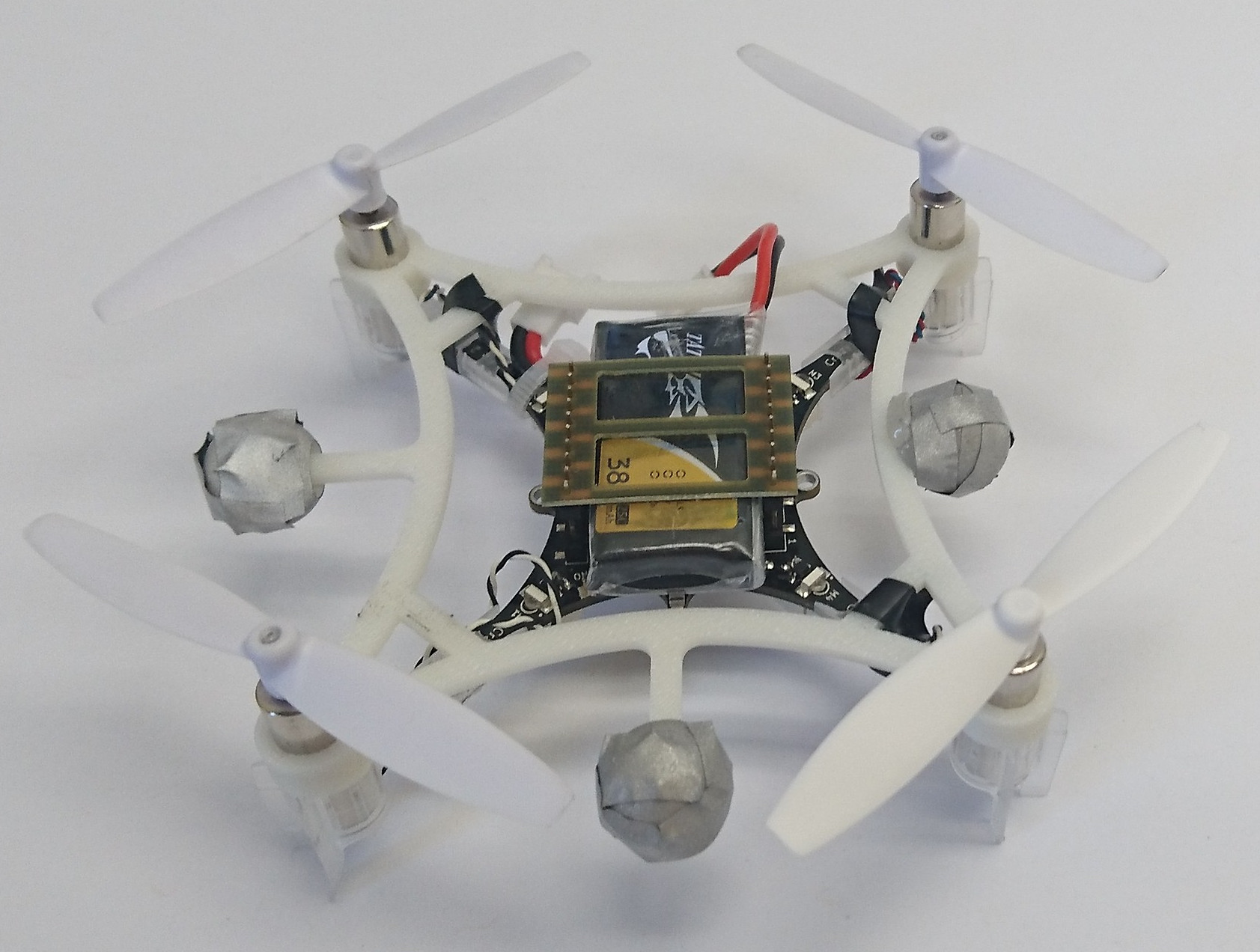}
  \caption{
  The quadcopter used in the experiments, measuring approximately 160mm from propeller tip to opposing propeller tip.
  }
  \label{figExpQuad}
\end{figure}

\subsection{Taking off with large yaw errors}
\begin{figure}
  \centering
  \includegraphics{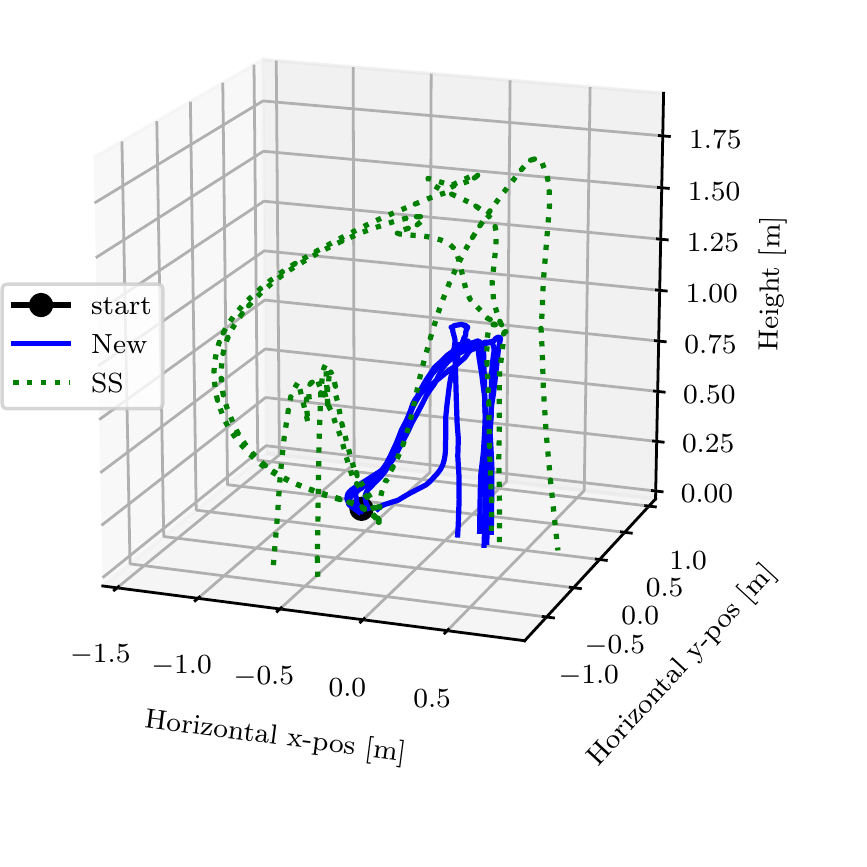}
  \caption{
  Position traces (for multiple experiments) for a quadcopter starting at the black dot, and commanded to fly to a position 0.5m horizontally along the $x$ direction, and at a height of 1.5m. 
  The quadcopter starts with a large yaw error.
  The variations in trajectories are due to noise in the system. 
  }
  \label{figExpYaw}
\end{figure}

The rotational symmetry of typical multicopters makes it easy for an operator to place a multicopter with a large initial yaw error, as can be seen in \figref{figExpQuad}. 
To illustrate the practical effects of such an error, a quadcopter was commanded to take off, and fly to a setpoint at height 1.5m, at a horizontal distance of approximately 0.5m away from the take-off position. 
The quadcopter, however, was initialized with a large yaw error (approximately 177$^\circ$).
The position traces for five experiments each with the new controller, and the skew-symmetric controller, are shown in \figref{figExpYaw}. 
As expected from \secref{secPerfAdvantageTiltPrior}, the skew-symmetric controller performs very poorly, and in all cases the vehicle meanders a substantial distance from the target point before reaching it (if it does not collide with a wall first). 
The proposed controller does not exhibit this behavior.

\subsection{Recovery from large initial errors}
\begin{figure}
  \centering
  \includegraphics{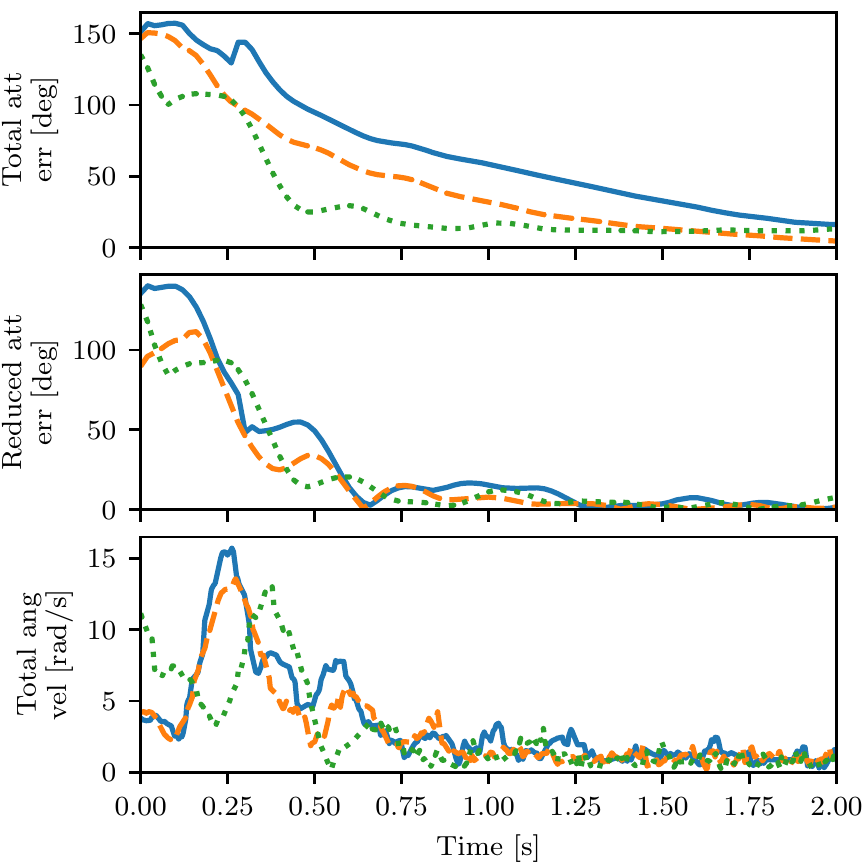}
  \caption{
  Experimental results for a vehicle recovering from large initial disturbances using the proposed controller, after being thrown into the air by a user. 
  The control action starts just after time 0, and each line style identifies the same experiment in the three plots.
  The attitude error is as estimated from motion capture data, and the angular velocity as measured by a rate gyroscope.
  }
  \label{figExpLargeDisturbance}
\end{figure}

To demonstrate the performance of the proposed controller when recovering from large disturbances, a series of experiments were performed where the quadcopter was thrown aloft by a user, and the controller only activated once the vehicle exceeds a certain height threshold. 
The results of these experiments are shown in \figref{figExpLargeDisturbance} -- from the figure it can be seen that the reduced attitude error is rapidly controlled to zero, while the total attitude error may decay much slower. 
This matches the intuition from the prior numerical examples.

\section{Discussion and conclusion}
\label{secConclusion}

In this paper we have compared four attitude controllers for multicopters, three from the literature and one novel controller. 
All perform identically to first order. 
However, it was shown that the controller with angular acceleration proportional to the skew-symmetric part of the rotation matrix presents a safety concern, as large attitude errors may persist for an arbitrary period, if the total angular error is near $180^\circ$ (note that this behavior exists over a substantial neighborhood near that point).
Using an angular acceleration proportional to the rotation vector representation of the attitude error is shown to be preferable, as no such dangers exist. 
Moreover, any pure initial rotation will decay like a damped second-order system, making for more intuitive behavior. 

The decomposition of the attitude error into a tilt and yaw component allows for controllers that prioritize the vehicle's thrust direction, and therefore potentially more quickly converge that part of the vehicle attitude which dominates translational motion.
We presented the quaternion-based tilt-prioritizing controller of \cite{brescianini2013nonlinear}, wherein the angular acceleration is proportional to the sin of half the error angles.
This controller did not pose any of the safety concerns that the skew-symmetric controller has.

Inspired by this, we also presented a novel controller, which prioritizes the vehicle tilt, but does so with angular acceleration proportional to the angles. 
The controller is posed using rotation axes and angles, making for an intuitive description, and stability is shown with a relatively simple Lyapunov function. 
Specifically, the control action is a combination of the control action resulting from controlling proportionally to the rotation vector and proportionally to the only tilt error.
Though it is inspired by, the quaternion-based tilt-prioritizing controller, the behavior is distinct therefrom, and closed-loop performance compares favorably with the quaternion-based controller. 
The novel controller, furthermore, continues to perform well if the tilt is not prioritized much over the yaw angle, specifically converging to the rotation-vector based control in the limit as the control weight on the yaw direction converges to that of the tilt direction. 
Experimental results validate the controller performance under realistic conditions. 

Thus, we recommend the application of the novel controller for multicopters, especially where robustness to large disturbances is desired. 
The controller out-performs standard controllers, specifically it does not suffer from the poor convergence from large attitudes that the skew-symmetric controller shows; it outperforms the rotation-vector based controller in most circumstances; and the attitude error converges faster (using potentially less agressive inputs) that the tilt-prioritizing quaternion-based controller.

\section*{Acknowledgements}
This research was supported by funding from the Powley foundation.
We also thank Koushil Sreenath and Dario Brescianini for their valuable inputs. 

\bibliographystyle{plainnat}
\bibliography{bibliography}

%\bibliographystyle{IEEEtran}
%\bibliography{bibliography}

\end{document}